\documentclass[Times1COL]{WileyNJDv5} 
\usepackage[style=apa]{biblatex} 
\usepackage{tcolorbox}
\usepackage{float}
\usepackage{placeins} 
\usepackage{csquotes}
\articletype{Original Article}
\received{Date Month Year}
\revised{Date Month Year}
\accepted{Date Month Year}
\journal{BJET}
\volume{00}
\copyyear{2024}
\startpage{1}

\graphicspath{{figures/}}

\begin{document}
\title{Using Generative AI and Multi-Agents to Provide Automatic Feedback}
  \author[1]{Shuchen Guo}

 \author[2,3,4]{Ehsan Latif}

\author[2,3,4]{Yifan Zhou}

\author[5]{Xuan Huang}

\author[2,3,4]{Xiaoming Zhai}

\authormark{Guo \textsc{et al.}}
\titlemark{Using Generative AI and Multi-Agents to Provide Automatic Feedback}

\address[1]{\orgdiv{School of Teacher Education}, \orgname{Nanjing Normal University}, \orgaddress{\city{Nanjing}, \state{Jiangsu}, \country{China}}}

\address[2]{\orgdiv{AI4STEM Education Center}, \orgname{University of Georgia}, \orgaddress{\city{Athens}, \state{Athens}, \country{USA}}}

\address[3]{\orgdiv{National GENIUS Center}, \orgname{University of Georgia}, \orgaddress{\city{Athens}, \state{Athens}, \country{USA}}}

\address[4]{\orgdiv{Department of Mathematics, Science, and Social Studies Education}, \orgname{University of Georgia}, \orgaddress{\city{Athens}, \state{Athens}, \country{USA}}}

\address[5]{\orgname{Beijing Institute of Education}, \orgaddress{\city{Beijing}, \country{China}}}

\corres{Corresponding author Xiaoming Zhai. \email{xiaoming.zhai@uga.edu}}


\fundingInfo{This study is supported by the National Science Foundation (\#2101104) and Institute of Education Sciences (\#R305C240010). Any opinions, findings, conclusions, or recommendations expressed in this material are those of the author(s) and do not necessarily reflect the views of the NSF or IES.}

\abstract[Abstract]{This study investigates the use of generative AI and multi-agent systems to provide automatic feedback in educational contexts, particularly for student constructed responses in science assessments. The research addresses a key gap in the field by exploring how multi-agent systems, called AutoFeedback, can improve the quality of GenAI-generated feedback, overcoming known issues such as over-praise and over-inference that are common in single-agent large language models (LLMs). The study developed a multi-agent system consisting of two AI agents: one for generating feedback and another for validating and refining it. The system was tested on a dataset of 240 student responses, and its performance was compared to that of a single-agent LLM. Results showed that AutoFeedback significantly reduced the occurrence of over-praise and over-inference errors, providing more accurate and pedagogically sound feedback. The findings suggest that multi-agent systems can offer a more reliable solution for generating automated feedback in educational settings, highlighting their potential for scalable and personalized learning support. These results have important implications for educators and researchers seeking to leverage AI in formative assessments, offering a pathway to more effective feedback mechanisms that enhance student learning outcomes.}




\maketitle

\renewcommand\thefootnote{}
\footnotetext{\textbf{Abbreviations:} LLM, Large Language Model; GenAI, Generative Artificial Intelligence}

\renewcommand\thefootnote{\fnsymbol{footnote}}
\setcounter{footnote}{1}


\begin{tcolorbox}[colback=white!10!white, colframe=gray!60!gray,  
                  width=\textwidth, 
                  boxrule=1.5mm, 
                 ]
    
    \textbf{\large{Practitioner notes}} \\
    \\
    What is already known about this topic?
    \vspace{-1em}
    \begin{itemize}
        \item Automatic feedback is essential for supporting personalized learning when effectively provided.
        \item GenAI-generated automatic feedback has advantages, but drawbacks like over-praise and over-inference need to be addressed.
        \item Multi-agents have the potential for conducting complex educational tasks.
    \end{itemize}
    What this paper adds?
        \vspace{-1em}
    \begin{itemize}
        \item This research developed multi-agents for improving the quality of GenAI-generated feedback in terms of addressing over-praise and over-inference problems.
        \item The results show that the multi-agents outperform single agents by significantly reducing the occurrence of over-praise and over-inference. 
    \end{itemize}
    Implications for practice and/or policy
    \vspace{-1em}
    \begin{itemize}
        \item This study proposed multi-agents as a promising way to improve the quality of GenAI-generated automatic feedback.
    \end{itemize}

\end{tcolorbox}

\section{Introduction}
Automatic feedback strengthens teachers' capacity by offering tailored, real-time information to support student learning, with the potential to improve students' learning experience and outcomes \parencite{hahn2021systematic}. Automatic feedback is especially valuable in online learning environments \parencite{chauhan2014massive, cavalcanti2021automatic} and classroom formative assessment practices, as it addresses the challenges that instructors have limited time to provide individualized feedback simultaneously to students with diverse needs. Research has consistently reported the positive effects of automatic feedback, including increased student engagement, improved learning outcomes, and reduced teachers’ bias \parencite{hahn2021systematic,kochmar2020automated, razzaq2020effect}. With the development of Artificial Intelligence (AI) techniques, particularly generative AI (GenAI) such as ChatGPT, many argue that teachers can leverage these tools to automatically generate natural-sounding and context-specific feedback for more complex tasks, which holds the potential to significantly advance student learning  \parencite{guo2024artificial,du2024harnessing,fokides2024comparing,guo2024resist}. 


However, GenAI-powered automatic feedback also demonstrated limitations\parencite{jansen2024comparing,scarlatos2024improving}. Among the many drawbacks of GenAI-generated feedback, over-praise and over-inference can be detrimental to student learning (Authors, under review). Over-praise occurs when the feedback is excessively positive, even if the students' responses are incorrect or meaningless. Over-inference refers to a misalignment between the student's actual performance and the intepretation in the feedback, making the feedback inaccurately matched to the student's work. These two issues can convey misleading information that undermines students' learning, decreasing learning motivation and fostering inauthentic self-esteem, ultimately impairing opportunities to improve learning \parencite{lee2017understanding,irvin2one}. 

 To address the issues of over-praise and over-inference, this study developed a multi-agent feedback system (AutoFeedback) to generate quality automatic feedback for students’ written responses in science assessments, with one agent to generate and another to examine and revise feedback. The two agents in the multi-agent system interact with each other, playing different roles and serving varying functions. Compared to the single agent, multi-agents are more powerful for addressing complex problems due to the scalability, robustness, reliability as a whole, and specialization and heterogeneity per agent. Therefore, this approach holds the potential to help automatically provide better feedback. We tested AutoFeedback's performance with 240 students' constructed responses and compared the feedback quality with that of a single agent. The study answers two questions: 1) How frequently does a single GenAI agent generate feedback with over-praise and over-inference? 2) To what degree does the multi-agent system AutoFeedback improve feedback in terms of reducing over-praise and over-inference?

\section{AI-based Automatic Feedback for Learning}
\subsection{Significance of Learning Feedback}
Feedback is one of the most essential components to support student learning, which can assist students in assessing their learning progress, identifying learning gaps, and improving self-regulated skills. Effective feedback can produce more learning gains and increase students' satisfaction \parencite{black1998assessment}. 
Due to the importance of feedback, research has highlighted the characteristics of quality feedback. For example, Hattie and Timperley proposed a framework articulating the key features of feedback \parencite{hattie2007power}. They summarise that effective feedback should answer three questions—“Where am I going?”,  “How am I going?” and “Where to next?” The three questions highlight that feedback has to not only address students' current performance but also point to where students are heading (i.e., learning goals) and how they could improve their learning (i.e., learning support). Additionally, researcher further conceptualized feedback as dialogical and stressed seven valuable features for effective feedback: understandable, selective, specific, timely, contextualized, non-judgemental, balanced, forward-looking, transferable, and personal \parencite{nicol2014monologue}. Despite this recommendation for sound and good practice of feedback, it is challenging to implement them in an authentic educational context where students' needs are diverse and regular personal contact between teacher and student is difficult. Thus, automatic feedback generation has been gaining increasing attention for its potential to provide real-time, personalized feedback to support learning. 

\subsection{Automatic Feedback Generation}
Traditionally, automatic feedback has been provided based on comparing student answers and the desired solution \parencite{cavalcanti2021automatic}. For tasks that already have structured or definite answers, the strategy can help students work through certain exercises in disciplines. In this case, potential responses are already recorded in the system, allowing for immediate feedback through comparison. However, a significant limitation of this method is its inability to handle unexpected or emergent responses, thereby discouraging innovative thinking
\parencite{hahn2021systematic}. 
With the growing integration of AI in education, automatic feedback has evolved to support more complex assessment tasks, such as constructed responses and essays, through natural language processing (NLP) and machine learning (ML) techniques \parencite{hahn2021systematic,cavalcanti2019analysis}. 
These systems, trained on datasets consisting of student responses and human ratings, can recognize patterns and evaluate new texts on similar tasks \parencite{jansen2024comparing}. Empirical research shows that NLP/ML can provide quality feedback by examining and evaluating substantial volumes of text through syntactic and semantic analysis algorithms \parencite{sailer2023adaptive,bernius2022machine}. However, building such systems is time- and resource-intensive due to the extensive data required for training and the associated costs. Moreover, teachers and educational researchers with limited knowledge of programming or NLP/ML techniques face barriers to using these systems effectively \parencite{jansen2024comparing,lee2024applying}.

Recently, GenAI, such as ChatGPT, have emerged as promising tools for automatic feedback, offering the potential to overcome these challenges. GenAI not only possesses a higher degree of linguistic comprehension and generation abilities but also can complete challenging cognitive tasks in ways that are more approachable and user-friendly for the public, without requiring extensive datasets typically needed for traditional NLP/ML systems \parencite{jansen2024comparing,lee2024applying}. Research has demonstrated the potential of ChatGPT for automatic feedback generation. Research explored the potential of GenAI for automatically identifying student errors in scientific inquiry, providing a foundation for productive, personalized feedback \parencite{bewersdorff2023assessing}. Based on GPT-3.5 and GPT-4, they developed an AI system that accurately identifies fundamental errors in experimentation. Research investigated ChatGPT’s performance in generating feedback on students’ argumentative writing in English using prompt engineering and revealed that ChatGPT-generated feedback addressed significantly more aspects of student work compared to teacher feedback, though teachers expressed both positive and negative opinions on its features \parencite{guo2024resist}. The authors ultimately recommend that teachers collaborate with ChatGPT to provide feedback on student writing. Despite the potential, the quality of GenAI-powered feedback requires further attention, as it still has undeniable flaws that need to be resolved \parencite{jansen2024comparing}.

\subsection{Quality of GenAI-generated Automatic Feedback: Over-Praise and Over-Inference}
Empirical studies have shown that GenAI-generated automatic feedback has notable drawbacks. For example, five aspects have been identified that GenAI struggles with when generating feedback: correctness, revealing answer, suggestion, diagnostic value, and positive tone \cite{scarlatos2024improving}. Research further compared GenAI-generated feedback with human-generated feedback and found that humans outperformed GenAI in providing clearer directions for improvement, greater accuracy, prioritizing essential features, and using a supportive tone \cite{steiss2024comparing}. Fokides and Peristeraki analyzed the efficacy of ChatGPT in generating feedback for primary school students’ short essays in both the English and Greek languages. They found that ChatGPT's performance was language-sensitive, as it provided less satisfying feedback compared to educators in Greek \parencite{fokides2024comparing}. 

Apart from the potential and issues mentioned above, over-praise and over-inference are also usually observed in GenAI-powered automatic feedback. Praise refers to a positive evaluation of another’s work, performance, or attributes \parencite{kanouse1981semantics}. Over-praise occurs when feedback provides overly positive encouragement that does not align with the student's actual performance. While praise can motivate and reward students, enhancing their self-esteem, studies found that general, indiscriminate, or overly positive praise is unhelpful \parencite{gan2021teacher}. This can be particularly problematic for low-achieving students, where excessive praise may have a negative impact \parencite{black1998assessment}. Since effective feedback should be both accurate and constructive, over-praise can hinder students' learning by creating false impressions of their performance. Over-inference occurs when feedback draws conclusions that cannot be reasonably inferred from the student's response. Research has recognized that large language models (LLMs) struggle with processing the linguistic diversity and subtle nuances of human language, which can lead to over-inference and negatively impact feedback quality \cite{fuchs2023exploring}. In education, however, only when feedback is based on an assessment of students' actual responses can it provide helpful, constructive information and suggestions for future learning and metacognitive development. Automatic feedback with over-inference can mislead students, resulting in misunderstandings of the content and inaccurate self-evaluations. 

Despite the known drawbacks in GenAI-generated automatic feedback, research to evaluate and improve feedback quality remains limited \parencite{cavalcanti2021automatic,jansen2024comparing}. Most studies compared automatic feedback with
human-generated feedback, even though human-generated feedback can also have limitations. Existing research has primarily focused on feedback generation, particularly for English learners, with fewer studies addressing methods to improve feedback based on identified issues. Therefore, in this study, we focused on two key issues, over-praise and over-inference, and contributed by proposing AutoFeedback, a multi-agent, as a promising solution for improving GenAI-generated feedback on constructive responses.

\section{Multi-Agents for Automatic Feedback}


\subsection{Multi-Agents}

An agent in AI refers to an autonomous entity powered by language models that can plan, reason, and execute tasks to achieve a specific goal. Agents operate by receiving input from the environment, interpreting it through reasoning, and then performing actions accordingly. This iterative process allows agents to execute tasks that require problem-solving, decision-making, and even interaction with external tools \parencite{Li2023tools}. Multi-agent systems involve multiple autonomous agents collaborating to achieve a unified goal \parencite{Guo2023multiagent, Qian2023collaboration}. Each agent in this framework can be specialized for a specific role \parencite{Huang2023hallucination}. Multi-agent systems extend the functionality of single agents by incorporating multiple agents that can work together, either independently or in collaboration. 

Unlike single-agent architectures, multi-agent systems allow for division of labor, parallel processing, and enhanced robustness through feedback and interaction between agents\parencite{Hong2023metagpt}. Having multiple agents ensures robustness, as errors can be cross-checked and reduced, and it allows for better management of the complexities of tasks. By distributing tasks across agents, a multi-agent system increases the accuracy and precision of response, providing more relevant, targeted information to users. The evolution of multi-agent systems has been marked by significant advancements, particularly with the integration of LLMs. Initially, research focused on training autonomous agents in isolated environments, limiting their ability to make human-like decisions. With the advent of LLMs, researchers began exploring their potential in collaborative settings, leading to the emergence of frameworks that enable multiple agents to work together effectively.

In the field of education, multi-agent systems have been employed for various purposes \parencite{Tang2023eduagent}. It is particularly useful in complex educational processes in which different aspects should be considered. For example, to promote adaptive learning, multi-agent systems have been developed where each agent can focus on a different aspect of the learning process, such as identifying knowledge gaps, adjusting the difficulty of content, or providing personalized feedback. This helps cater to different learning styles and paces, making learning more efficient and student-centered. MEDCO \parencite{Wei2024medco} is a multi-agent copilot system for medical education that simulates real-world training environments by incorporating roles such as a patient, doctor, and radiologist, fostering interactive, multi-disciplinary learning and enhancing students’ question-asking and collaborative skills in a virtual setting \parencite{Williams2023medco}. SimClass is a multi-agent classroom simulation framework using LLMs to emulate traditional classroom interactions, incorporating real user participation and enabling collaborative, role-based teaching dynamics to enhance the learning experience \parencite{Du2023simclass,Zhang2024simclass}. Major findings indicate that multi-agent systems can enhance learning experiences and respond to individual student needs, offering a more personalized and effective approach to personalized learning \parencite{Chen2023dynamiclearning}.

\begin{figure}[htp]
    \centering
    \includegraphics[width=0.9\linewidth]{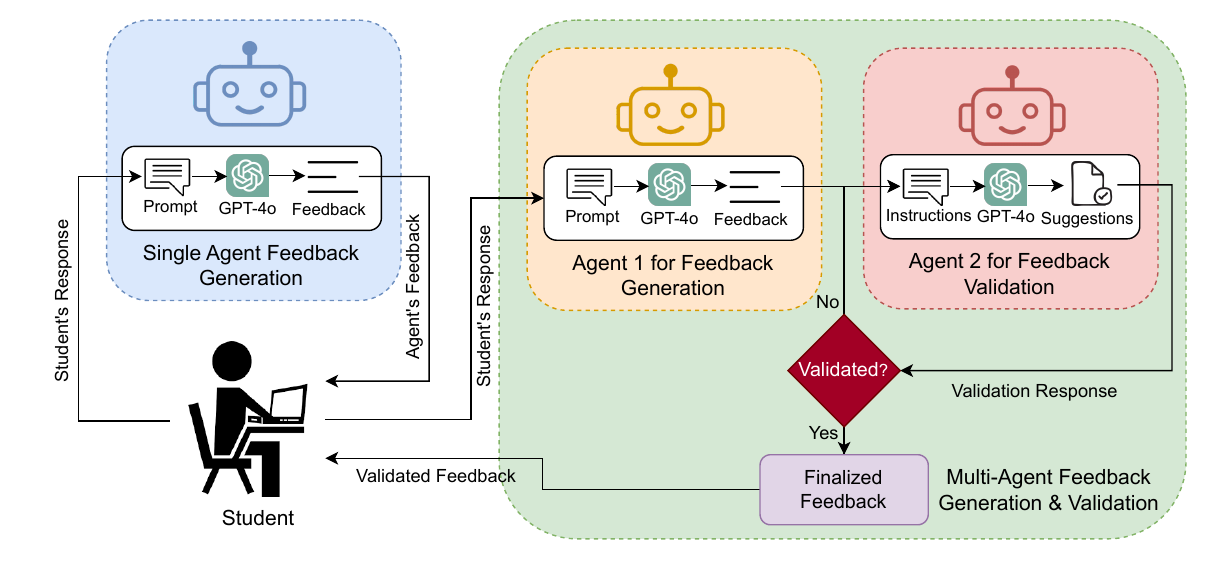}
    \caption{Architecture for multi-agent vs single agent system for automatic feedback generation.}
    \label{fig:single_vs_multi_agent_system}
\end{figure}

\subsection{Multi-Agent for Automatic Feedback}
\label{multi_agent_system}

Automatic feedback systems powered by AI technologies \parencite{wongvorachan2022artificial}, particularly those incorporating GenAI and multi-agent systems, have the potential to significantly enhance the speed, personalization, and scalability of educational feedback \parencite{yesilyurt2023ai}. These systems offer immediate and individualized evaluations of student work, addressing key challenges such as handling large-scale assessments. The multi-agent framework (See Fig.~\ref{fig:single_vs_multi_agent_system}) ensures that feedback is continuously evaluated and refined, providing an additional layer of quality control to mitigate issues like over-praise or over-inference. The design of a multi-agent system for automatic feedback generation involves multiple AI-driven agents working in tandem to enhance the quality, speed, and personalization of feedback in educational settings. As depicted in Figure \ref{fig:single_vs_multi_agent_system}, this study employs an architecture consisting of several specialized agents that collaboratively produce and validate feedback for student responses. The system, called AutoFeedback, is composed of the following key components:

\textbf{Feedback Generation Agent (Agent 1)}: This agent generates the initial feedback based on a student's submission. It uses an LLM like GPT-4, fine-tuned for educational contexts. The agent generates feedback tailored to the student's response, incorporating elements such as assessment goals, student performance, and areas for improvement. Predefined rubrics or templates can guide the feedback generation to ensure consistency and alignment with instructional goals. The model generates feedback in natural language, offering insights into the strengths and weaknesses of the response. Parameters such as verbosity, tone, and focus areas (e.g., content accuracy, grammar, or creativity) can be configured depending on the instructional requirements.

\textbf{Feedback Validation Agent (Agent 2)}: After the initial feedback is generated, the validation agent evaluates the quality of the feedback. This agent is designed to minimize the risks of over-praise or over-inference by checking the generated feedback against predefined validation rules. These rules may include criteria such as correctness, fairness, relevance, and completeness. 

The workflow of the \textit{AutoFeedback} system can be broken down into the following steps:
\begin{enumerate}
\item \textbf{Student Submission}: The process begins when students submit their work, such as an essay, problem solution, or other response.
\item \textbf{Initial Feedback Generation}: Agent 1 processes the student's submission and generates preliminary feedback using an LLM. The feedback is based on predefined templates, rubrics, or other instructional guidelines.
\item \textbf{Feedback Validation and Revision}: The generated feedback is passed to Agent 2, which evaluates it against pedagogical and quality control criteria. This involves both semantic and structural analysis of the feedback.
\item \textbf{Refinement (if needed)}: If the validation agent identifies any issues, it sends the feedback back to the generation agent with suggestions for improvement. The feedback is regenerated or refined according to these suggestions.
\item \textbf{Final Approval and Delivery}: Once the modification agent approves the feedback, it is finalized and delivered to the student.
\end{enumerate}

The framework underpinning this system, demonstrates how generative AI models, when integrated into multi-agent infrastructures, can be effectively deployed to address the complexities of educational feedback. Combining generative AI's capacity for nuanced language understanding with AutoFeedback's ability to review and improve outputs systematically, this hybrid approach offers a promising solution to the complexities of educational feedback. It can potentially improve both the effectiveness and efficiency of assessment processes, ultimately benefiting teachers and students in large-scale educational settings.


\section{Methods}
\subsection{Item and Student Responses}
The dataset comes from existing study, which asked middle school students to construct and describe scientific models for science phenomena \parencite{zhai2022applying}. The task was designed to examine whether students meet the NGSS performance expectation, MS-PS1-4 (MS: Middle School, PS: Physical Sciences), specifically \textit{develop a model that explains how particle motion changes when thermal energy is transferred to or from a substance without changing state}. The stem of the item shows a context with three dishes of water at cool, room, and hot temperatures. Shwan dropped a red-coated chocolate candy into each dish. Students can then watch a video showing what happens in the three dishes. The item asks students to construct a model to show and describe what is happening to explain the phenomena. To answer the question, students are expected to use a model to describe how the transfer of thermal energy affects particle motion and/or temperature. The responses were scored using a rubric that categorized students as either "Beginning" or "Proficient."
A total of 845 student responses were collected and scored by human raters. We created a balanced testing dataset for automatic feedback generation by randomly selecting 120 (50\%) "Beginning" responses and 120 (50\%) "Proficient" responses. The balanced dataset helped to minimize errors, potential bias, and performance inflation. Additionally, we randomly sampled 30 student responses to make a balanced pilot dataset for developing prompts. 

\subsection{Experiments}
The experimental setup involved simulating a realistic feedback generation scenario in which both single- and multi-agent system was tasked with processing the responses to generate high-quality feedback. For the single agent, we employed Agent 1, whose construction and development process was described below, to generate initial feedback. This stage mirrored how traditional LLM-based systems might produce feedback, making it possible to directly compare the single-agent approach to the multi-agent method. For AutoFeedback, the multi-agents, Agent 2 was added to evaluate and revise the feedback produced by Agent 1, identifying instances where the feedback might include over-praise or over-inference. If such issues were identified, Agent 2 further revised the feedback to address the problem. In each case, the multi-agent system processed the student response, first generating feedback using Agent 1 and then passing the generated feedback to Agent 2 for review and refinement (details given in Sec.~\ref{multi_agent_system}). Unlike traditional LLM feedback systems, the multi-agent architecture (See Fig.~\ref{fig:single_vs_multi_agent_system}) enabled a second layer of quality control, ensuring that feedback not only aligned with the instructional goals but also avoided the common pitfalls of over-praise and over-inference. The experiments allowed us to compare the multi-agent system to a baseline approach, where only a single LLM-generated feedback without any subsequent validation.

We have used GPT-4o as the base model for both agents, and they communicate in sequence to get the desired validated feedback.
The experiments were run on a local system by using open APIs\footnote{\url{https://platform.openai.com/docs/api-reference/introduction}} capable of handling large-scale processing, ensuring that the system could manage the volume of responses without performance degradation. Each assessment item was handled independently, and the responses were processed in parallel to simulate real-world educational scenarios where multiple students submit their work simultaneously. The experiments tested the system’s ability to handle diverse student submissions while maintaining a high standard of feedback quality.

\subsubsection{Agent 1: Feedback Generator}
Following the WRVRT prompting approach \parencite{lee2024applying}, we developed the prompts for agents. First, researchers wrote a prompt for the automatic feedback. Then, to ensure face validity, one educational assessment research expert and one AI research expert reviewed the prompt and provided suggestions for revising. Next, with 30 randomly selected student responses, we validated and revised the prompt again. The iterative WRVRT is completed until the prompts reach saturation. Finally, the researcher ran the test data with the prompts.

For Agent 1, which is responsible for automatic feedback generation, the prompt combined five components. (See Appendix Fig.~\ref{fig:prompt_agent1} for complete sample prompt)

\textit{Role} instructs GPT’s role as a middle school science teacher doing formative assessment and has received student responses. 

\textit{Task} describes the aim of the task is to suggest the student with feedback to help improve their science learning based on their response to the item. For the prompts that provide GPT with \textit{ItemRubRes}, \textit{Context and }\textit{CRITERIA FOR FEEDBACK}, a sentence that instructs GPT to refer to these elements concatenated with \textit{Task}.

\textit{Item Rubrics }first provides GPT with the item prompt and scoring rubric. The scoring rubric lists three scoring components. Then, student response is provided.

\textit{Context} provides GPT with the basic teaching and learning context of the item. It describes the subject, school level, and the assessment aim of the item. Then, 3D learning goals, including core scientific concepts, crosscutting concepts, and science and engineering practices associated with the item, were also described.

\textit{Criteria for Feedback} describes the quality of effective feedback. Firstly, according to Hattie and Timperley (2007), the feedback should include three main parts: the aim of the item, student performance in terms of strength and areas for improvement, and suggestions for further learning. At the same time, we ask GPT to generate feedback that acquires characteristics like understandable to middle school students, balanced, actionable, encouraging, and so on. These features are concluded based on the suggestions proposed by Nicol (2007).

\subsubsection{Agent 2: Feedback Validation and Modifier}
Agent 2 is in charge of detecting over-praise and over-inference issues in the feedback generated by Agent 1 and revising it if needed. The prompt of Agent 2 included some same components as Agent 1, such as \textit{ ItemRubRes} and \textit{Context}. Besides, it includes some different components as presented below. (See Appendix Fig.~\ref{fig:prompt_agent2} for complete sample prompt)

\textit{Role} instructs GPT’s role as a middle school science teacher doing formative assessment, who has received student responses as well as feedback provided by ChatGPT. 

\textit{Task} describes the aim of the task is to evaluate if the feedback should be revised or not by providing reasons. Revise the feedback if needed.  For the prompts that provide GPT with the following components, a sentence that instructs GPT to refer to these elements concatenated with \textit{Task}.

\textit{Feedback} provides GPT with the automatic feedback generated by agent 1.

\textit{Possible problems of feedback} describes over-praise and over-inference problems, providing simple examples.

\subsection{Data Analysis}
After receiving the automatic feedback responses, two assistant professors researching science education evaluated the feedback according to the definition of over-praise and over-inference. Over-praise happens when automatic feedback is provided over positive encouragement when a student's actual performance is not good enough to match it. Over-inference was coded when automatic feedback drew some conclusion that could not be directly inferred from the student response. Besides, there are cases being labeled as over-praise and over-inference at the same time when both issues were found in the feedback.

For the entire coding task, the two raters coded 30\% cases separately, and the inter-rater reliability reached 100\% after a discussion. Then, the remaining cases were coded independently by one rater. The human analysis results were used to compare the performance of a single LLM agent and a multi-agent.

\section{Findings}

\subsection{Feedback Generated by Single Agent}

This study found that 15.42\% of the automatically generated feedback provided by a single GenAI agent was labeled as over-praise, while 27.20\% was labeled as over-inference. Additionally, 9.58\% of the feedback exhibited both over-praise and over-inference. The coding results are presented in Table 1.

For issues of over-praise, Fig. 2 illustrates a typical case (Stu 40912) where the student's response consisted of meaningless characters. Despite this, the automatic feedback stated, "It's great that you attempted to respond!" and "You're on the right path to understanding these concepts," which are overly positive given that the student's answer had no relevance to the task.

\begin{figure}[h]
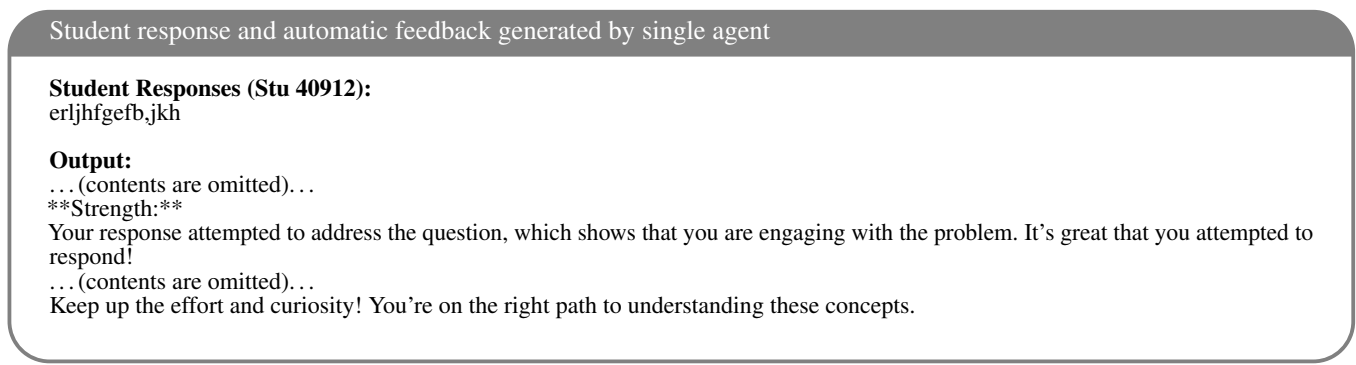
 
\begin{tcolorbox}[colback=white, colframe=gray, width=\textwidth, arc=5mm, auto outer arc, boxrule=0.5mm, title=Student response and automatic feedback generated by single agent, fontupper=\scriptsize]
    \textbf{Student Responses (Stu 40912):} \\
    erljhfgefb,jkh \\
    \\
    \textbf{Output:}\\
    …(contents are omitted)…\\
**Strength:**\\
Your response attempted to address the question, which shows that you are engaging with the problem. It's great that you attempted to respond!\\
…(contents are omitted)…\\
Keep up the effort and curiosity! You're on the right path to understanding these concepts.\\
\end{tcolorbox}
\caption{Feedback with over-praise problem}
\label{fig:prompt_agent1}
\end{figure}

Fig. 3 shows an example of an over-inference. In this case (Stu 36652), the student's answer did not mention anything related to "temperature." However, the automatic feedback inferred that the student was 'beginning to consider what happens when the candy dissolves in water at different temperatures,' which was an unwarranted evaluation.

\begin{figure}[h]
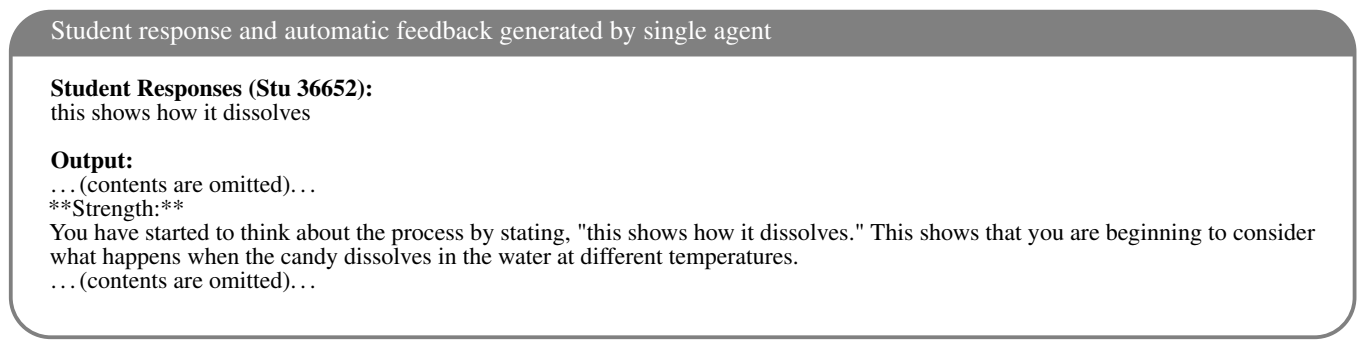
 
\begin{tcolorbox}[colback=white, colframe=gray, width=\textwidth, arc=5mm, auto outer arc, boxrule=0.5mm, title=Student response and automatic feedback generated by single agent, fontupper=\scriptsize]
    \textbf{Student Responses (Stu 36652):} \\
   this shows how it dissolves \\
    \\
    \textbf{Output:}\\
    …(contents are omitted)…\\
**Strength:**\\
You have started to think about the process by stating, "this shows how it dissolves." This shows that you are beginning to consider what happens when the candy dissolves in the water at different temperatures.\\
…(contents are omitted)…\\
\end{tcolorbox}
\caption{Feedback with over-inference problem}
\label{fig:prompt_agent1}
\end{figure}

The case shown in Fig. 4 was labeled as both over-praise and over-inference. In the case (Stu 19042), the student's response described how the red particles dissolved at different temperatures but did not connect this phenomenon to the core concept being assessed. However, the feedback inferred that the student 'understands that thermal energy affects particle movement' and included praise such as "Well done!" This feedback misjudged the student's performance and provided commendation for that,  clearly exemplifying both over-praise and over-inference.

\begin{figure}[h]
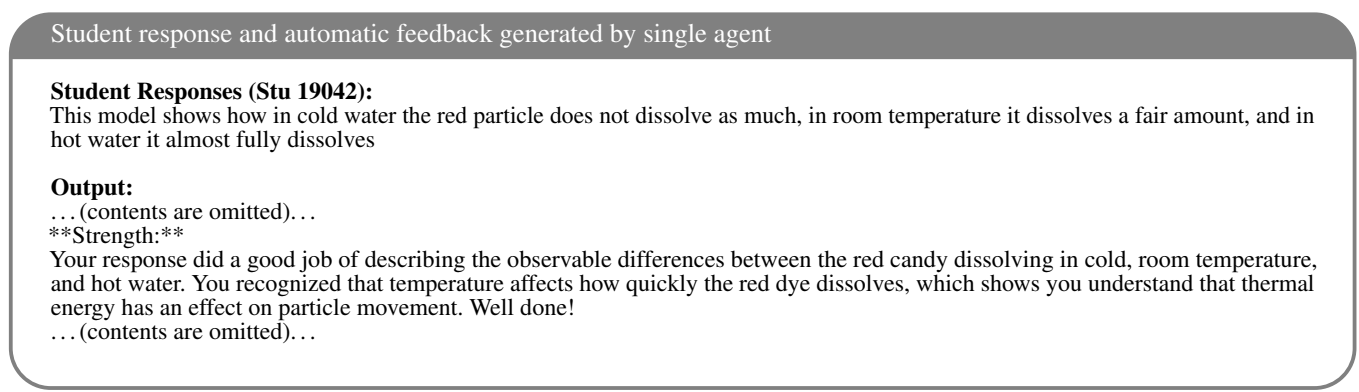
 
\begin{tcolorbox}[colback=white, colframe=gray, width=\textwidth, arc=5mm, auto outer arc, boxrule=0.5mm, title=Student response and automatic feedback generated by single agent, fontupper=\scriptsize]
    \textbf{Student Responses (Stu 19042):} \\
  This model shows how in cold water the red particle does not dissolve as much, in room temperature it dissolves a fair amount, and in hot water it almost fully dissolves\\
    \\
    \textbf{Output:}\\
    …(contents are omitted)…\\
**Strength:**\\
Your response did a good job of describing the observable differences between the red candy dissolving in cold, room temperature, and hot water. You recognized that temperature affects how quickly the red dye dissolves, which shows you understand that thermal energy has an effect on particle movement. Well done!\\
…(contents are omitted)…\\
\end{tcolorbox}
\caption{Feedback with over-praise and over-inference problems}
\label{fig:prompt_agent1}
\end{figure}

\subsection{Feedback Generated by AutoFeedback}

To address the issues of over-praise and over-inference, we applied AutoFeedback to provide feedback to the same group of students' performance. Table 1 presents the results of feedback generated by the single-agent and multi-agent systems in terms of over-praise and over-inference issues, along with the chi-square test results for performance comparison. The chi-square test revealed that the AutoFeedback generated significantly less feedback with over-praise and/or over-inference problems (p < 0.001), indicating that the GenAI multi-agent significantly outperformed the GenAI single-agent system, resulting in a higher proportion of more qualified automatic feedback.

\begin{table}[h]
    \centering 
    \caption{\textit{Comparsion of feedback generated by single and multi-agent LLM}} 
    \begin{tabular}{p{2cm} p{2cm} p{2cm} p{4cm}} 
        \hline
        Agent & Over praise & Over inference & Over praise and over inference \\ 
        \hline
        Single agent & 37/15.42 & 68/27.20 & 23/9.58 \\ 
        Multi-agent & 3/1.25 & 17/7.08 & 2/0.83 \\ 
        Chi-square & 31.527 & 37.185 & 18.609 \\ 
        p & 0.000 & 0.000 & 0.000 \\  
        \hline
    \end{tabular}
    \label{tab:key_findings}
\end{table}

For the issue of over-praise, we found a decrease in the percentage of over-praise feedback from 15.42\% to 1.2\%. AutoFeedback first identified the over-praise issue and then revised the feedback accordingly. For instance, in the over-praise case shown in Fig. 2 (Stu 40912), Agent 2 flagged the problem by noting, "The feedback praises the attempt ("Your response attempted to address the question"), but given that the student's response ("erljhfgefb,jkh") is gibberish, this is over-praised." The feedback was then revised to better align with the student's actual response, as shown in Fig. 5.

\begin{figure}[h] 
\begin{tcolorbox}[colback=white, colframe=gray, width=\textwidth, arc=5mm, auto outer arc, boxrule=0.5mm, title=Student response and automatic feedback generated by multi-agents, fontupper=\scriptsize]
    \textbf{Student Responses (Stu 40912):} \\
    erljhfgefb,jkh \\
    \\
    \textbf{Output:}\\
    …(contents are omitted)…\\
**Strength:**\\
Your response submission indicates that you were present to complete the task, which is a necessary step. However, there is room for significant improvement to address the question fully.\\
…(contents are omitted)…\\
Understanding how to create meaningful sentences that clearly describe scientific observations is key. Keep practicing these steps to improve your understanding of thermal energy and particle motion.\\
Keep trying, and don’t hesitate to ask for help if you need it!\\
\end{tcolorbox}
\caption{Revised feedback solving over-praise problem}
\label{fig:prompt_agent1}
\end{figure}

For over-inference, the percentage decreased to 7.08\% with the use of AutoFeedback. Similarly, in the case shown in Fig. 3 (Stu 36652), the multi-agent response pointed out, "The feedback infers the student is beginning to consider what happens when the candy dissolves in water at different temperatures. There is no evidence in the response to support this inference". Fig. 6 displays the revised feedback, which reflects the student's description of the phenomena more accurately, and provides appropriate suggestions for improving the response.

\begin{figure}[h] 
\begin{tcolorbox}[colback=white, colframe=gray, width=\textwidth, arc=5mm, auto outer arc, boxrule=0.5mm, title=Student response and automatic feedback generated by multi-agents, fontupper=\scriptsize]
    \textbf{Student Responses (Stu 36652):} \\
this shows how it dissolves
\\
    \\
    \textbf{Output:}\\
    …(contents are omitted)…\\
**Strength:**\\
- It's good that you are attempting to describe a process. Starting with any observation is a positive step.
\\
**Area for improvement:**\\
- Your response needs more detail to show your understanding of how thermal energy affects particle motion. Specifically, you need to explain how the particles in the water move faster when heated and slower when cooled. Additionally, your explanation should be written in clear, complete sentences.
\\
 …(contents are omitted)…\\
\end{tcolorbox}
\caption{Revised feedback solving over-inference problem}
\label{fig:prompt_agent1}
\end{figure}

For feedback that contained both over-praise and over-inference problems, the percentage decreased to 0.83\% with the use of AutoFeedback. In the case shown in Fig. 4 (Stu 19042), AutoFeedback identified both problems, stating that "The feedback seems too positive given the student did not fully address how particle motion changes with the transfer of thermal energy, which is crucial based on the scoring rubric." and "The feedback infers understanding that wasn't clearly demonstrated in the student response. For example, it states the student recognized that temperature affects how quickly the red dye dissolves when the response doesn't explicitly talk about particle motion." Fig. 7 shows the revised feedback, which resolved both issues and more accurately reflected the student's performance.

\begin{figure}[h] 
\begin{tcolorbox}[colback=white, colframe=gray, width=\textwidth, arc=5mm, auto outer arc, boxrule=0.5mm, title=Student response and automatic feedback generated by multi-agents, fontupper=\scriptsize]
    \textbf{Student Responses (Stu 19042):} \\
This model shows how in cold water the red particle does not dissolve as much, in room temperature it dissolves a fair amount, and in hot water it almost fully dissolves.
\\
    \\
    \textbf{Output:}\\
    …(contents are omitted)…\\
**Strength:**\\
- Your response describes the observable differences between the red candy dissolving in cold, room temperature, and hot water. You noted that the candy dissolves to different extents based on temperature, indicating an understanding that temperature plays a role.
\\
\end{tcolorbox}
\caption{Revised feedback solving over-praise and over-inference problem}
\label{fig:prompt_agent1}
\end{figure}

However, we have to admit that even when using multi-agents for generating automatic feedback, some responses still suffer from issues of over-praise and over-inference. In some cases, these issues were detected but not fully revised. Sometimes  Agent 2 can provide an incorrect evaluation of the feedback, leading to unsatisfactory revisions.

\section{Conclusions and Discussion }\label{sec5}
AI-powered automatic feedback has been viewed as useful venues to support students' personalized learning, however, issues such as over-praise or over-inferences can desrupt students' learning and raising signtificant concerns. To address these issues, we developed a GenAI multi-agent, AutoFeedback, to generate quality automatic feedback. To examine its usefulness, we compared AutoFeedback with a single GenAI agent in providing feedback to students’ written explanations of scientific phenomena. The study found that although both systems can provide feedback, AutoFeedback significantly outperformed single agents in reducing over-praise and over-inference problems. By using AutoFeedback, the occurrence of over-praise and over-inference decreased by 14.17\% and 20.21\%, respectively, compared to a single agent. For feedback containing both over-praise and over-inference (9.58\%), AutoFeedback reduced the percentage by 8.75\%. These improvements highlight the potential of using multi-agents as a promising way to provide more accurate and effective automatic feedback, which is vital for better assessing student performance and progression, enabling more informed decisions for future learning and teaching. To the best of our knowledge, this is the first study exploring the potential of a multi-agent GenAI system for improving the quality of automatic feedback, with one agent generating feedback and the other responsible for identifying issues and revising accordingly.

Despite the great potential of GenAI-powered automatic feedback, this study found a significant portion of feedback includes either over-praise or over-inferences. For feedback to be effective, it must provide accurate evaluation and constructive suggestions for student learning in a positive yet realistic manner. While positive feedback is important, praise should align with students' actual performance, as over-praise can be detrimental. Irvin et al. also found that students prefer and value teachers who praise only high-quality work rather than those who praise all work equally highly \parencite{irvin2one}.  Lee et al. found that from children's perspectives, over-praise is associated with poorer school performance and higher depression compared to accurately praising \parencite{lee2017understanding}. Over-praise can also decrease motivation, making students see no need to strive for further improvement. Studies also found that GenAI-generated tends to include significantly more praise than that provided by teachers \parencite{guo2024resist}. 

As for over-inference, it essentially provides inaccurate evaluations of student performance, which can mislead learning and distort self-perception. The issue may stem from the accuracy and linguistic limitations of LLMs. Studies have emphasized the two limitations as challenges for integrating GenAI in education. Due to these drawbacks, GenAI might struggle to understand the nuances and complexities of human language, leading to misunderstandings and incorrect responses \parencite{fuchs2023exploring}. 

Despite of over-praise and over-inference, some studies have uncovered other drawbacks of GenAI-generated feedback by comparing it to human feedback. Steiss et al. compared ChatGPT-generated feedback with human-generated feedback using five specific criteria drawn from existing educational research: the degree to which the feedback was criteria-based, provided clear directions for improvement, was accurate, prioritized essential writing features, and used a supportive tone \parencite{steiss2024comparing}. They found that human evaluators provided higher-quality feedback than ChatGPT in all the dimensions except criteria-based feedback. Jansen et al. evaluated the effectiveness of feedback on argumentative writing provided by ChatGPT-3.5 and human experts, using a 10-point Likert-scale to collect student teachers' opinions \parencite{jansen2024comparing}. The results showed that the participants rated ChatGPT-generated feedback as useful for 59\%, compared to 88\% for feedback provided by human experts. Based on all the evidence above, we believe that the quality of GenAI-generated feedback still needs further improvement. Until then, both teachers and students should be careful and critical in using automatic feedback.


Previous studies have uncovered some drawbacks of GenAI-powered automatic feedback when leveraging a single GenAI agent to conduct the task. Similarly, in this study, we observed obvious issues of over-praise and over-inference when using a single GenAI agent to generate automatic feedback. However, based on the results of AutoFeedback, we found that multi-agents can significantly improve the quality of GenAI-generated automatic feedback by addressing over-praise and over-inference issues. Multi-Agents are widely used in the fields where cooperative effort is required to fulfill the purpose of the end product \parencite{viswanathan2022enhancement}. In this study, the final product is quality automatic feedback free of over-praise and over-inference. The two agents in AutoFeedback work collaboratively to generate, review, and revise feedback. With each agent specializing in its specific task, AutoFeedback can address over-praise and over-inference effectively, providing significantly more accurate feedback than a single agent. 
Researchers have already recognized the potential of multi-agent systems for managing the complexity of the educational domain and improving efficiency and speed compared to other systems \parencite{viswanathan2022enhancement}.
For example, many studies have employed multi-agents to enhance online educational systems for adaptive learning \parencite{viswanathan2022enhancement,alexandru2015enhanced}, recommender system \parencite{amane2021multi}, and intelligent tutoring system \parencite{azevedo2012metatutor}.
Lmati (2022) \cite{Lmati2022} proposed a new multi-agent approach for generating feedback based on multiple-choice questions. However, in the study, the feedback was not advice or comments but rather multiple-choice questions. Lmati used three agents in the study: a learner agent representing the learner, a feedback agent who sends the multiple-choice questions to the learners who made mistakes, and a controller agent who updated the question base based on learner feedback. While this system facilitated interaction between the learner and the platform, it did not provide automatic feedback as generally understood. 
To our knowledge, this study is the first to leverage  GenAI multi-agents to generate and improve automatic feedback for constructive responses in science education. The results indicate that multi-agents can be a promising solution for improving the quality of GenAI-powered feedback. Potentially, as Agent 2 serves the function of detecting and revising feedback, similar strategies could be employed to address other common issues found in automatic feedback.

\section{Limitations and Future Study}\label{sec5}
While the GenAI multi-agents significantly improve the quality of feedback by reducing issues of over-praise and over-inference, future research should continue exploring ways to further enhance the effectiveness of automatic feedback. Although AutoFeedback developed in this study significantly lowered the occurrence of over-praise and over-inference, Agent 2 did not always detect these problems with complete accuracy. At the same time, not all of the issues identified by Agent 2 were fully resolved in the final feedback. 
Moreover, beyond the over-praise and over-inference issues highlighted in this study, other concerns regarding feedback quality should also draw attention. For example, one of the advantages of automatic feedback is to provide customized feedback for personalized learning. However, we found that the suggestions for further learning provided by ChatGPT tended to be similar across students. In the study, we only provide ChatGPT with student responses to the only item. With limited information, it is hard for ChatGPT to generate truly personalized learning strategies. It should also be acknowledged that even human educators find it difficult to offer highly personalized learning suggestions for each student. 
Considering these limitations,  future research should further explore advanced prompt engineering and multi-agent design to continuously improve the quality of automatic feedback. Expanding the range of assessment items used in these systems is also necessary for further exploration. Beyond theoretical considerations of feedback quality, it is also crucial to investigate the real-world impact of automatic feedback on student learning in authentic educational settings.



\newpage
\appendix
\renewcommand{\thefigure}{A\arabic{figure}}
\begin{figure}[ht!] 
\begin{tcolorbox}[colback=white, colframe=gray, width=\textwidth, arc=5mm, auto outer arc, boxrule=0.5mm, title=Prompt, fontupper=\scriptsize]
    \textbf{Role:} \\
    You are a middle school science teacher who is conducting a formative assessment in science teaching. You provided open-ended items to students and received responses. You are going to provide feedback to students to help improve their science learning. \\
    \\
    \textbf{TASK:}\\
    You will receive <<ITEM>>, <<SCORING RUBRIC 1>> and <<STUDENT RESPONSE>>.\\ With consideration <<TEACHING AND LEARNING CONTEXT>>, <<3D LEARNING GOAL>> and << CRITERIA FOR FEEDBACK>>, suggest the student with feedback to help improve their science learning. \\
    \\
    \texttt{<<ITEM>>}\\ 
    \textit{Problem Statement:}
     Shwan had 3 dishes of water at room temperature. She cooled one dish, causing thermal energy to transfer from that dish to the surroundings. She kept the middle dish at room temperature. She transferred thermal energy into the third dish by heating it. Then Shwan dropped a red-coated chocolate candy into each dish.\\ 
     \textit{Question:} Use a model to describe how the transfer of thermal energy affects particle motion and/or temperature.\\
     \\
    \texttt{<<SCORING RUBRIC 1>>}\\
    -[Rule 1]: When thermal energy is transferred to the water (hotter), water and dye particles move faster. \\  
    -[Rule 2]: At a higher temperature, water and dye particles move faster. \\
    -[Rule 3]: The answer is a meaningful sentence.\\ 
    The level of the student will be ‘Proficient’ if the response includes ([Rule 1] OR [Rule 2]) AND [Rule 3]. The left will be ‘Beginning’.\\
    \\
    \texttt{<<STUDENT RESPONSE>>}\\
    …(contents are omitted)…\\
    \\
    \texttt{<<TEACHING AND LEARNING CONTEXT>> }\\
    - School level is middle school. The subject is Physics. \\
    - This task measures a student’s proficiency in the following: Develop a model that explains how particle motion changes when thermal energy is transferred to or from a substance without changing state. \\
    \\
    \texttt{<<3D LEARNING GOAL>>} \\
    -Core Concept:  liquids are made of molecules or inert atoms that are moving about relative to each other. Adding or removing thermal energy increases or decreases the kinetic energy of the particles.\\ 
    -Cross-cutting concept: Cause and effect relationships may be used to predict phenomena in natural or designed systems. \\
    -Science and engineering practice: Develop a model to predict and/or describe phenomena. \\
    \\
    \texttt{<<CRITERIA FOR THE FEEDBACK>>}\\
    \textit{- Content:}  \\
    \textit{Aim of the item:}  (The feedback should first tell the student the aim of the item according to the learning goal) \\
    Your performance \\ 
    \textit{Strength:}  show where student did well and why according to <<3D LEARNING GOAL>> and <<SCORING RUBRIC 1>>.\\ 
    \textit{Area for improvement:} show where need improvement and why according to <<3D LEARNING GOAL>> and <<SCORING RUBRIC 1>>.\\ 
    \textit{Suggestions for further learning:}  (based on Area for improvement and <<3D LEARNING GOAL>>, provide suggestions for students to improve their learning. Name concrete activities students can do. If the student is ‘proficient’ level, suggest more advanced learning activities. \\

    \textit{- Structure:} \\ 
    The feedback needs to be presented in a constructive and facilitated manner. \\ 
    The feedback should be understandable to middle school students.\\  
    Try to be concise and detailed at the same time. No more than 300 words. \\
    The feedback needs to be encouraging. \\
    DO NOT directly say about the <<3D LEARNING GOAL>> or <<SCORING RUBRIC 1>> in the feedback. \\
    DO NOT directly tell the answer to the item in the Area for improvement and Suggestions for further learning.\\

    \vspace{0.5cm}
\end{tcolorbox}
\caption{Prompt for Agent1 for Sample Assessment Item}
\label{fig:prompt_agent1}
\end{figure}

\begin{figure}[!htbp] 
\begin{tcolorbox}[colback=white, colframe=gray, width=\textwidth, arc=5mm, auto outer arc, boxrule=0.5mm, title=Prompt, fontupper=\scriptsize]
    \textbf{Role:} \\
    You are a middle school science teacher who is conducting a formative assessment in science teaching. You provided open-ended items to students and received responses. You then received the feedback provided by ChatGPT 4o based on the students' response. You want to see whether the feedback is good enough to give to students and revise it if it is not good enough.\\
    \\
    \textbf{TASK:}\\
    You will receive <<ITEM>>, <<SCORING RUBRIC 1>>, <<STUDENT RESPONSE>> and <<FEEDBACK FROM AGENT1>>. With consideration <<TEACHING AND LEARNING CONTEXT>>, and << POSSIBLE PROBLEM OF FEEDBACK>>, proceed with the below tasks step by step:\\
    STEP 1. Evaluate if <<FEEDBACK FROM AGENT1>> should be revised or not. Give the reasons. \\
    STEP 2. If <<FEEDBACK FROM AGENT1>> do not need revision, say “The feedback now is good enough”. If <<FEEDBACK FROM AGENT1>> need revision, revise it with remaining the components including \textit{Aim of the Item}, \textit{Your Performance}, \textit{Strength}, \textit{Area for improvement}  and \textit{Suggestions for further learning}.
 \\
    \\
    \texttt{<<ITEM>>}\\ 
    \textit{Problem Statement:}
     Shwan had 3 dishes of water at room temperature. She cooled one dish, causing thermal energy to transfer from that dish to the surrounding. She kept the middle dish at room temperature. She transferred thermal energy into the third dish by heating it. Then Shwan dropped a red -coated chocolate candy into each dish.\\ 
     \textit{Question:} Use a model to describe how the transfer of thermal energy affects particle motion and/or temperature.\\
     \\
    \texttt{<<SCORING RUBRIC 1>>}\\
    -[Rule 1]: When thermal energy is transferred to the water (hotter), water and dye particles move faster. \\  
    -[Rule 2]: At the higher temperature, water and dye particles move faster. \\
    -[Rule 3]: The answer is a meaningful sentence.\\ 
    The level of student will be ‘Proficient’ if the response include ([Rule 1] OR [Rule 2]) AND [Rule 3]. The left will be ‘Beginning’.\\
    \\
    \texttt{<<STUDENT RESPONSE>> }\\
    …(contents are omitted)… \\
    \\
    \texttt{<<TEACHING AND LEARNING CONTEXT>> }\\
    - School level is middle school. The subject is Physics. \\
    - This task measures a student’s proficiency in the following: Develop a model that explains how particle motion changes when thermal energy is transferred to or from a substance without changing state. \\
    \\
    \texttt{<<3D LEARNING GOAL>>} \\
    -Core Concept:  liquids are made of molecules or inert atoms that are moving about relative to each other. Adding or removing thermal energy increases or decreases kinetic energy of the particles.\\ 
    -Cross-cutting concept: Cause and effect relationships may be used to predict phenomena in natural or designed systems. \\
    -Science and engineering practice: Develop a model to predict and/or describe phenomena. \\
    \\
    \texttt{<<FEEDBACK FROM AGENT1>>}\\
    …(contents are omitted)…
    \\
    
    \texttt{<<POSSIBLE PROBLEM OF FEEDBACK>>}\\
    \textit{Over-praise:} the feedback is too positive about the student’s response even if it does not. For example, say “You're doing great” or “You’re on the right path” when student are actually not.\\
    \textit{Over-inference:} There is a misalignment between the feedback and the student's actual performance. The feedback says something that cannot be inferred directly from the student’s actual response according to the scoring rubric and 3D learning goal. For example, <<FEEDBACK FROM AGENT1>> say words that are not in <<STUDENT RESPONSE>>.\\

\end{tcolorbox}
\caption{Prompt for Agent2 for Sample Assessment Item}
\label{fig:prompt_agent2}
\end{figure}

\newpage
\FloatBarrier
\printbibliography

@article{zhai2022applying,
  title={Applying machine learning to automatically assess scientific models},
  author={Zhai, Xiaoming and He, Peng and Krajcik, Joseph},
  journal={Journal of Research in Science Teaching},
  volume={59},
  number={10},
  pages={1765--1794},
  year={2022},
  publisher={Wiley Online Library}
}

@article{jansen2024comparing,
  title={Comparing Generative AI and Expert Feedback to Students’ Writing: Insights from Student Teachers},
  author={Thorben Jansen and Lars Höft and Luca Bahr and Johanna Fleckenstein and Jens Möller and Olaf Köller and Jennifer Meyer},
  journal={Psychologie in Erziehung und Unterricht},
  volume={71},
  number={2},
  pages={80--92},
  year={2024},
  publisher={Ernst Reinhardt Verlag},
  doi={10.2378/peu2024.art08d}
}

@article{lee2024applying,
  title={Applying large language models and chain-of-thought for automatic scoring},
  author={Lee, Gyeong-Geon and Latif, Ehsan and Wu, Xuansheng and Liu, Ninghao and Zhai, Xiaoming},
  journal={Computers and Education: Artificial Intelligence},
  volume={6},
  pages={100213},
  year={2024},
  publisher={Elsevier}
}

@article{du2024harnessing,
  title={Harnessing large language models to auto-evaluate the student project reports},
  author={Du, Haoze and Jia, Qinjin and Gehringer, Edward and Wang, Xianfang},
  journal={Computers and Education: Artificial Intelligence},
  volume={7},
  pages={100268},
  year={2024},
  publisher={Elsevier}
}

@inproceedings{fuchs2023exploring,
  title={Exploring the opportunities and challenges of NLP models in higher education: is Chat GPT a blessing or a curse?},
  author={Fuchs, Kevin},
  booktitle={Frontiers in Education},
  volume={8},
  pages={1166682},
  year={2023},
  organization={Frontiers Media SA}
}

@article{viswanathan2022enhancement,
  title={Enhancement of online education system by using a multi-agent approach},
  author={Viswanathan, Nethra and Meacham, Sofia and Adedoyin, Festus Fatai},
  journal={Computers and Education: Artificial Intelligence},
  volume={3},
  pages={100057},
  year={2022},
  publisher={Elsevier}
}

@article{alexandru2015enhanced,
  title={Enhanced education by using intelligent agents in multi-agent adaptive e-learning systems},
  author={Alexandru, Adriana and Tirziu, Eugenia and Tudora, Eleonora and Bica, Ovidiu},
  journal={Studies in Informatics and Control},
  volume={24},
  number={1},
  pages={13--22},
  year={2015}
}

@inproceedings{amane2021multi,
  title={A multi-agent and content-based course recommender system for university e-learning platforms},
  author={Amane, Meryem and Aissaoui, Karima and Berrada, Mohammed},
  booktitle={International Conference on Digital Technologies and Applications},
  pages={663--672},
  year={2021},
  organization={Springer}
}

@inproceedings{azevedo2012metatutor,
  title={MetaTutor: An intelligent multi-agent tutoring system designed to detect, track, model, and foster self-regulated learning},
  author={Azevedo, Roger and Bouchet, Fran{\c{c}}ois and Harley, Jason M and Feyzi-Behnagh, Reza and Trevors, Gregory and Duffy, Melissa and Taub, Michelle and Pacampara, Nicole and Agnew, Lauren and Griscom, Sophie and others},
  booktitle={The Fourth Workshop on Self-Regulated Learning in Educational Technologies (SRL\&ET)},
  year={2012}
}

@article{Lmati2022,
  author    = {Lmati, I.},
  title     = {New multi-agent approach for generating feedbacks based on Multiple Choice Questions},
  journal   = {International Journal for Innovation Education and Research},
  year      = {2022},
  volume    = {10},
  number    = {1},
  pages     = {209--220},
  doi       = {10.31686/ijier.vol10.iss1.3608},
  url       = {https://doi.org/10.31686/ijier.vol10.iss1.3608}
}

@article{steiss2024comparing,
  title={Comparing the quality of human and ChatGPT feedback of students’ writing},
  author={Steiss, Jacob and Tate, Tamara and Graham, Steve and Cruz, Jazmin and Hebert, Michael and Wang, Jiali and Moon, Youngsun and Tseng, Waverly and Warschauer, Mark and Olson, Carol Booth},
  journal={Learning and Instruction},
  volume={91},
  pages={101894},
  year={2024},
  publisher={Elsevier}
}

@inproceedings{kochmar2020automated,
  title={Automated personalized feedback improves learning gains in an intelligent tutoring system},
  author={Kochmar, Ekaterina and Vu, Dung Do and Belfer, Robert and Gupta, Varun and Serban, Iulian Vlad and Pineau, Joelle},
  booktitle={Artificial Intelligence in Education: 21st International Conference, AIED 2020, Ifrane, Morocco, July 6--10, 2020, Proceedings, Part II 21},
  pages={140--146},
  year={2020},
  organization={Springer}
}

@inproceedings{razzaq2020effect,
  title={Effect of immediate feedback on math achievement at the high school level},
  author={Razzaq, Renah and Ostrow, Korinn S and Heffernan, Neil T},
  booktitle={International Conference on Artificial Intelligence in Education},
  pages={263--267},
  year={2020},
  organization={Springer}
}

@article{guo2024artificial,
  title={Artificial intelligence in education research during 2013--2023: A review based on bibliometric analysis},
  author={Guo, S. and Zheng, Y. and Zhai, X.},
  journal={Education and Information Technologies},
  pages={1--23},
  year={2024},
  publisher={Springer}
}

@inproceedings{scarlatos2024improving,
  title={Improving the validity of automatically generated feedback via reinforcement learning},
  author={Scarlatos, Alexander and Smith, Digory and Woodhead, Simon and Lan, Andrew},
  booktitle={International Conference on Artificial Intelligence in Education},
  pages={280--294},
  year={2024},
  organization={Springer}
}

@article{lee2017understanding,
  title={Understanding when parental praise leads to optimal child outcomes: Role of perceived praise accuracy},
  author={Lee, Hae In and Kim, Young-Hoon and Kesebir, Pelin and Han, Da Eun},
  journal={Social Psychological and Personality Science},
  volume={8},
  number={6},
  pages={679--688},
  year={2017},
  publisher={Sage Publications Sage CA: Los Angeles, CA}
}

@article{irvin2one,
  title={“This one’s great! That one’s okay.”: Investigating the role of selective vs. indiscriminate praise on children’s learning behaviors},
  author={Irvin, Molly Kennedy and Asaba, Mika and Stegall, Jessa and Frank, Michael and Gweon, Hyowon},
  journal={The undergraduate research journal of psychology at UCLA},
  volume={8},
  pages={50--82},
  year={2021},
  publisher={Springer}
}

@article{Guo2023multiagent,
  author = "Guo, Taicheng and Chen, Xiuying and Wang, Yaqi and Chang, Ruidi and Pei, Shichao and Chawla, Nitesh V and Wiest, Olaf and Zhang, Xiangliang",
  title = "Large Language Model based Multi-Agents: A Survey of Progress and Challenges",
  journal = "arXiv preprint arXiv:2402.01680v2",
  year = "2024"
}

@article{Qian2023collaboration,
  author = "Qian, Jiamin and Liu, Shuang and Xiao, Zihan",
  title = "Meta Programming for Multi-Agent Collaborative Framework",
  journal = "arXiv preprint arXiv:2308.00352",
  year = "2023"
}

@article{Li2023tools,
  author = "Li, Fei and Sun, Jian and Zhang, Qian",
  title = "API-Bank: A Comprehensive Benchmark for Tool-Augmented LLMs",
  journal = "arXiv preprint arXiv:2305.18365",
  year = "2023"
}

@article{Hong2023metagpt,
  author = "Hong, Sirui and Zheng, Xiawu and others",
  title = "MetaGPT: A Multi-Agent Framework for Collaborative Development",
  journal = "arXiv preprint arXiv:2308.00352",
  year = "2023"
}

@article{Tang2023eduagent,
  author = "Tang, Xinyuan and others",
  title = "EduAgent: A Multi-Agent Framework for Educational Applications",
  journal = "arXiv preprint arXiv:2311.00368",
  year = "2023"
}

@article{Du2023simclass,
  author = "Du, Yan and others",
  title = "Simulating Classroom Interactions with Multi-Agent Systems",
  journal = "arXiv preprint arXiv:2312.11254",
  year = "2023"
}

@article{Williams2023medco,
  author = "Williams, Emma and Li, Chao",
  title = "MEDCO: A Multi-Agent Copilot System for Medical Education",
  journal = "arXiv preprint arXiv:2310.20195",
  year = "2023"
}

@article{Chen2023dynamiclearning,
  author = "Chen, Fang and Zhao, Li",
  title = "Dynamic Learning Experiences in Multi-Agent Systems for Education",
  journal = "arXiv preprint arXiv:2311.03289",
  year = "2023"
}

@article{Huang2023hallucination,
  author = "Huang, Lei and Yu, Weijiang and others",
  title = "A Survey on Hallucination in Large Language Models: Principles, Taxonomy, Challenges, and Open Questions",
  journal = "arXiv preprint arXiv:2311.05232",
  year = "2023",
  note = "\url{https://arxiv.org/abs/2311.05232}"
}

@article{Wei2024medco,
  author = "Wei, Hao and Qiu, Jianing and Yu, Haibao and Yuan, Wu",
  title = "MEDCO: Medical Education Copilots Based on A Multi-Agent Framework",
  journal = "arXiv preprint arXiv:2408.12496",
  year = "2024",
  url = "\url{https://doi.org/10.48550/arXiv.2408.12496}",
  note = "ECCV 2024 Workshop"
}

@article{Zhang2024simclass,
  author = "Zhang, Zheyuan and Zhang-Li, Daniel and Yu, Jifan and Gong, Linlu and Zhou, Jinchang and Liu, Zhiyuan and Hou, Lei and Li, Juanzi",
  title = "Simulating Classroom Education with LLM-Empowered Agents",
  journal = "arXiv preprint arXiv:2406.19226",
  year = "2024",
  url = "\url{https://doi.org/10.48550/arXiv.2406.19226}"
}

@article{gan2021teacher,
  title={Teacher feedback practices, student feedback motivation, and feedback behavior: how are they associated with learning outcomes?},
  author={Gan, Zhengdong and An, Zhujun and Liu, Fulan},
  journal={Frontiers in psychology},
  volume={12},
  pages={697045},
  year={2021},
  publisher={Frontiers Media SA}
}

@article{kanouse1981semantics,
  title={The semantics of praise},
  author={Kanouse, David E and Gumpert, Peter and Canavan-Gumpert, Donnah},
  journal={New directions in attribution research},
  volume={3},
  pages={97--115},
  year={1981}
}

@article{bewersdorff2023assessing,
  title={Assessing student errors in experimentation using artificial intelligence and large language models: A comparative study with human raters},
  author={Bewersdorff, Arne and Se{\ss}ler, Kathrin and Baur, Armin and Kasneci, Enkelejda and Nerdel, Claudia},
  journal={Computers and Education: Artificial Intelligence},
  volume={5},
  pages={100177},
  year={2023},
  publisher={Elsevier}
}

@article{bernius2022machine,
  title={Machine learning based feedback on textual student answers in large courses},
  author={Bernius, Jan Philip and Krusche, Stephan and Bruegge, Bernd},
  journal={Computers and Education: Artificial Intelligence},
  volume={3},
  pages={100081},
  year={2022},
  publisher={Elsevier}
}

@article{sailer2023adaptive,
  title={Adaptive feedback from artificial neural networks facilitates pre-service teachers’ diagnostic reasoning in simulation-based learning},
  author={Sailer, Michael and Bauer, Elisabeth and Hofmann, Riikka and Kiesewetter, Jan and Glas, Julia and Gurevych, Iryna and Fischer, Frank},
  journal={Learning and Instruction},
  volume={83},
  pages={101620},
  year={2023},
  publisher={Elsevier}
}

@inproceedings{cavalcanti2019analysis,
  title={An analysis of the use of good feedback practices in online learning courses},
  author={Cavalcanti, Anderson Pinheiro and de Mello, Rafael Ferreira Leite and Rolim, Vitor and Andr{\'e}, M{\'a}verick and Freitas, Fred and Ga{\v{s}}evic, Dragan},
  booktitle={2019 IEEE 19th international conference on advanced learning technologies (ICALT)},
  volume={2161},
  pages={153--157},
  year={2019},
  organization={IEEE}
}

@incollection{nicol2014monologue,
  title={From monologue to dialogue: improving written feedback processes in mass higher education},
  author={Nicol, David},
  booktitle={Approaches to assessment that enhance learning in higher education},
  pages={11--27},
  year={2014},
  publisher={Routledge}
}

@article{hattie2007power,
  title={The power of feedback},
  author={Hattie, John and Timperley, Helen},
  journal={Review of educational research},
  volume={77},
  number={1},
  pages={81--112},
  year={2007},
  publisher={Sage Publications Sage CA: Thousand Oaks, CA}
}

@article{black1998assessment,
  title={Assessment and classroom learning},
  author={Black, Paul and Wiliam, Dylan},
  journal={Assessment in Education: principles, policy \& practice},
  volume={5},
  number={1},
  pages={7--74},
  year={1998},
  publisher={Taylor \& Francis}
}

@article{fokides2024comparing,
  title={Comparing ChatGPT's correction and feedback comments with that of educators in the context of primary students' short essays written in English and Greek},
  author={Fokides, Emmanuel and Peristeraki, Eirini},
  journal={Education and Information Technologies},
  pages={1--45},
  year={2024},
  publisher={Springer}
}

@article{guo2024resist,
  title={To resist it or to embrace it? Examining ChatGPT’s potential to support teacher feedback in EFL writing},
  author={Guo, Kai and Wang, Deliang},
  journal={Education and Information Technologies},
  volume={29},
  number={7},
  pages={8435--8463},
  year={2024},
  publisher={Springer}
}

@article{hahn2021systematic,
  title={A systematic review of the effects of automatic scoring and automatic feedback in educational settings},
  author={Hahn, Marcelo Guerra and Navarro, Silvia Margarita Baldiris and Valent{\'\i}n, Luis De La Fuente and Burgos, Daniel},
  journal={IEEE Access},
  volume={9},
  pages={108190--108198},
  year={2021},
  publisher={IEEE}
}

@article{chauhan2014massive,
  title={Massive open online courses (MOOCS): Emerging trends in assessment and accreditation},
  author={Chauhan, Amit},
  journal={Digital Education Review},
  number={25},
  pages={7--17},
  year={2014},
  publisher={Digital Education Observatory (OED)}
}

@article{cavalcanti2021automatic,
  title={Automatic feedback in online learning environments: A systematic literature review},
  author={Cavalcanti, Anderson Pinheiro and Barbosa, Arthur and Carvalho, Ruan and Freitas, Fred and Tsai, Yi-Shan and Ga{\v{s}}evi{\'c}, Dragan and Mello, Rafael Ferreira},
  journal={Computers and Education: Artificial Intelligence},
  volume={2},
  pages={100027},
  year={2021},
  publisher={Elsevier}
}

@article{wongvorachan2022artificial,
  title={Artificial intelligence: Transforming the future of feedback in education},
  author={Wongvorachan, Tarid and Lai, Ka Wing and Bulut, Okan and Tsai, Yi-Shan and Chen, Guanliang},
  journal={Journal of Applied Testing Technology},
  pages={95--116},
  year={2022}
}

@incollection{yesilyurt2023ai,
  title={AI-enabled assessment and feedback mechanisms for language learning: Transforming pedagogy and learner experience},
  author={Yesilyurt, Yusuf Emre},
  booktitle={Transforming the Language Teaching Experience in the Age of AI},
  pages={25--43},
  year={2023},
  publisher={IGI Global}
}
\end{document}